\documentclass[journal]{IEEEtran}

\usepackage{hyperref}       
\usepackage{url}            
\usepackage{graphicx}  
\usepackage{amsfonts}
\usepackage{amsmath}
\usepackage{algorithm}
\usepackage{algpseudocode}
\usepackage{mathabx}
\usepackage{amssymb}
\usepackage{hyperref}

\usepackage[switch]{lineno}

\DeclareMathOperator*{\argmax}{argmax}
\usepackage[numbers]{natbib}
\begin{document}

\title{Enabling Humans to Plan Inspection Paths Using a Virtual Reality Interface}

\author{Boris~Bogaerts,
        Seppe~Sels,
        Steve~Vanlanduit
        and~Rudi~Penne
\thanks{B. Bogaerts, S. Sels, S. Vanlanduit and R. Penne are with the Faculty of applied engineering, University of Antwerp, Belgium,
 e-mail: boris.bogaerts@uantwerpen.be.}
}


\maketitle

\begin{abstract}
In this work, we investigate whether humans can manually generate high-quality robot paths for optical inspections. Typically, automated algorithms are used to solve the inspection planning problem. The use of automated algorithms implies that specialized knowledge from users is needed to set up the algorithm. We aim to replace this need for specialized experience, by entrusting a non-expert human user with the planning task. We augment this user with intuitive visualizations and interactions in virtual reality. To investigate if humans can generate high-quality inspection paths, we perform a user study in which users from different experience categories, generate inspection paths with the proposed virtual reality interface. From our study, it can be concluded that users without experience can generate high-quality inspection paths: The median inspection quality of user generated paths ranged between 66-81\% of the quality of a state-of-the-art automated algorithm on various inspection planning scenarios. We noticed however, a sizable variation in the performance of users, which is a result of some typical user behaviors. These behaviors are discussed, and possible solutions are provided. 
\end{abstract}

\begin{IEEEkeywords}
Human factors, Robotic inspection, Path planning, Non destructive testing, Virtual Reality
\end{IEEEkeywords}

\section{Introduction}
Robots are increasingly used for optical inspections as a means to provide motion to measurement devices (e.g. cameras). This motion is accurate, reliable and fast, which is important in driving down the cost of inspections. However, before a robot can automatically inspect an object it is necessary to find an efficient inspection path that the robot can follow. Traditionally, these inspection paths are generated either by experts, or by automated algorithms. However, to successfully use automated algorithms some pre-processing is required that requires experience in both robotics and optical inspection techniques \cite{bogaerts2019near}. Experience in optical inspection techniques is required to construct a quality function that models the fact that not all measurements are equal. These quality functions are typically generated using the physics of the specific measurement technique \cite{zhao2012automated,peeters2019optimized}. Robotics experience is necessary to construct a digital twin that can be accessed by the automated algorithm. This digital twin should contain the robot system, the measurement device, the object that is being inspected and the environment in which the inspection is performed. In the construction of this digital twin, a delicate trade-off between system fidelity and computational speed is important. On the one hand, optical robotic inspection systems are complex mechatronical systems that are nontrivial to construct, for which fidelity is important. But on the other hand finding automated inspection paths is a very challenging problem that requires many calls to the digital twin \cite{bogaerts2019near,englot2012sampling}. Overall, the potential of inspection planning is not realized in the inspection community, due to this requirement for specialized knowledge \cite{mineo2016robotic}.

The primary goal of this work is to construct a means for inexperienced users to generate high quality inspection paths. Our proposed system is a virtual reality interface in which users can generate inspection paths. This interface provides the user with intuitive visualizations and interactions that aim to replace the requirements for specialized knowledge. However, it is normal that more experience in the subject will lead to better inspection paths, so a second goal is to investigate the contribution of specialized knowledge towards generating high quality inspection paths. To do this, we perform a user study in which participants with different levels of experience in both robotics, and inspections generate inspection paths. Our final goal is to compare the quality of user generated inspection paths with automatically generated paths. This comparison will give an indication for which cases user input can be valuable in robotic inspection planning. This is important, especially since user input is traditionally not considered for generating inspection paths.

In \autoref{sec:RL} we will start with an overview of related literature. In \autoref{sec:theoryInspection} we will provide some necessary theoretical background on the inspection planning problem. In \autoref{sec:interface} we discuss details about our main contribution, namely a virtual reality interface. In \autoref{sec:experimentDesign} we will provide details on the experimental design of our user study, of which we will discuss the results in \autoref{sec:experimentResults}.

\section{Related work}
\label{sec:RL}
Robotic inspection planning is a diverse field with many applications. Inspection planning has already been applied to many optical non-destructive testing techniques with different robots. Some examples are:
\begin{itemize}
    \item Robot arm in thermography \cite{peeters2019optimized}, 3D reconstruction with optical cameras \cite{wenhardt2006information}
    \item 3D reconstruction with a laser line scanner on a coordinate measurement machine \cite{zhao2012automated}
    \item Drone based inspection with a thermal camera \cite{quater2014light}, optical camera \cite{bircher2016three}
    \item Autonomous underwater vehicle inspection with sonar \cite{englot2012sampling} 
\end{itemize}
All these seemingly different problems have been shown to be instances of the same general problem \cite{bogaerts2019near}. However all the aforementioned results were achieved with automated algorithms which require expertise to be used effectively. 

\citet{papachristos2016augmented} propose an Augmented Reality (AR) interface that allows users to control a drone during an inspection task. Their incentive for using human input is to deal with the lack of (or outdated) environment knowledge instead of improving the accessibility to inexperienced users.

Several articles try to combine the efforts of humans and automated algorithms in inspection path planning \cite{yi2014informative,reardon2018shaping}. \citet{reardon2018shaping} uses human interactions to bias the prior distribution in a sampling based algorithm. Users interact by coloring the solution space in areas that are likely to have good trajectories. Their results shows that using this kind of user interactions leads to higher quality inspection paths. \citet{yi2014informative} use an example path provided by a user, and a distance in which the automated algorithm can deviate from this path. This kind of user interaction significantly decreases the time needed to run the automated algorithm. However, instead of limiting the requirement of specialized input from users, these methods require additional input.

\citet{pintilie2013evaluation} compare the quality of user generated solutions with solutions generated by automated algorithms in a related problem, namely camera network design. They show that humans generate camera networks, with an intuitive interface, that are at the same quality level of those generated by automated algorithms. \citet{bogaerts2019interactive} designed a virtual reality interface that visualizes inspection quality to users. With this visualization inexperienced users are able to generate camera networks that are as good as automatically generated networks. The works \cite{pintilie2013evaluation,bogaerts2019interactive} are most closely related to the goals of this paper, especially as they focus on usability of the solution by users. However, we consider the inspection planning problem, a different, yet related problem. 

\citet{dry2006human} have shown that humans are able to generate solutions to the travelling salesman problem that are 11\% less efficient than the optimal solution. These results were obtained for graphs with 120 nodes. \citet{bogaerts2019interactive} on the other hand showed that humans are able to generate solutions as good as, or only slightly worse than the those of the optimal algorithm for submodular maximization problem with a cardinality constraint \cite{krause2008near}. The inspection planning problem combines submodular maximization with travelling salesman constraints \cite{bogaerts2019near}. In this work, we will also evaluate the quality of user generated solutions on the combination of submodular maximization, and the travelling salesman problem.

\section{Problem description}
\label{sec:theoryInspection}
In this section we will provide some essential background about the inspection planning problem that is needed to clarify some important points that we want to make\footnote{A VR360 video that summarizes this section is available online \url{https://youtu.be/7O629j58bTI}.}. We will also explain how the quality of user generated solutions can be fairly compared with each other, and how user generated solutions can be compared with solutions generated by an automated algorithm. In \autoref{sec:usability} we will discus the usability of our human based approach versus an automated algorithm based approach. This discussion will require some insight in the abstract problem of inspection planning.

\subsection{Inspection quality}
A central concept that is strictly necessary to compare different inspection paths is the notion of inspection quality functions. It is important to note that inspection quality is a concept that is viewed differently across different communities. In the robotics community it is often represented by the mutual information of some probabilistic distribution over measurement locations \cite{krause2008near}. In non-destructive testing, inspection quality is typically measured by custom functions that are dependent on the underlying physics of the measurement technique \cite{peeters2019optimized, zhao2012automated}. \citet{bogaerts2019near} have shown that although these quality functions are different, they can always be represented as submodular functions.

For any quality function it is essential that the object that needs to be inspected is represented by a finite amount of primitives. Without loss of generality, we assume that this object is provided as a mesh. In this article, we represent the object by a finite collection of points $M = \{m_1,...,m_n\}$. For practical purposes (i.e. the interactive quality visualization) we assume that these points are the vertices of the input mesh. In this work we further adopt a uniform distribution of points over the input mesh. We do this by publicly available re-meshing software \cite{jakob2015instant}. The measurement device can be positioned in a finite collection of view poses relative to the object that needs to be inspected $V=\{v_1,...,v_k\}$. Each view pose consists of a position and orientation of the measurement device. At each view pose only a subset of measurement locations are visible due to occlusion or the finite measurement volume of the measurement device.

\begin{figure*}
    \centering
    \includegraphics[width=\textwidth]{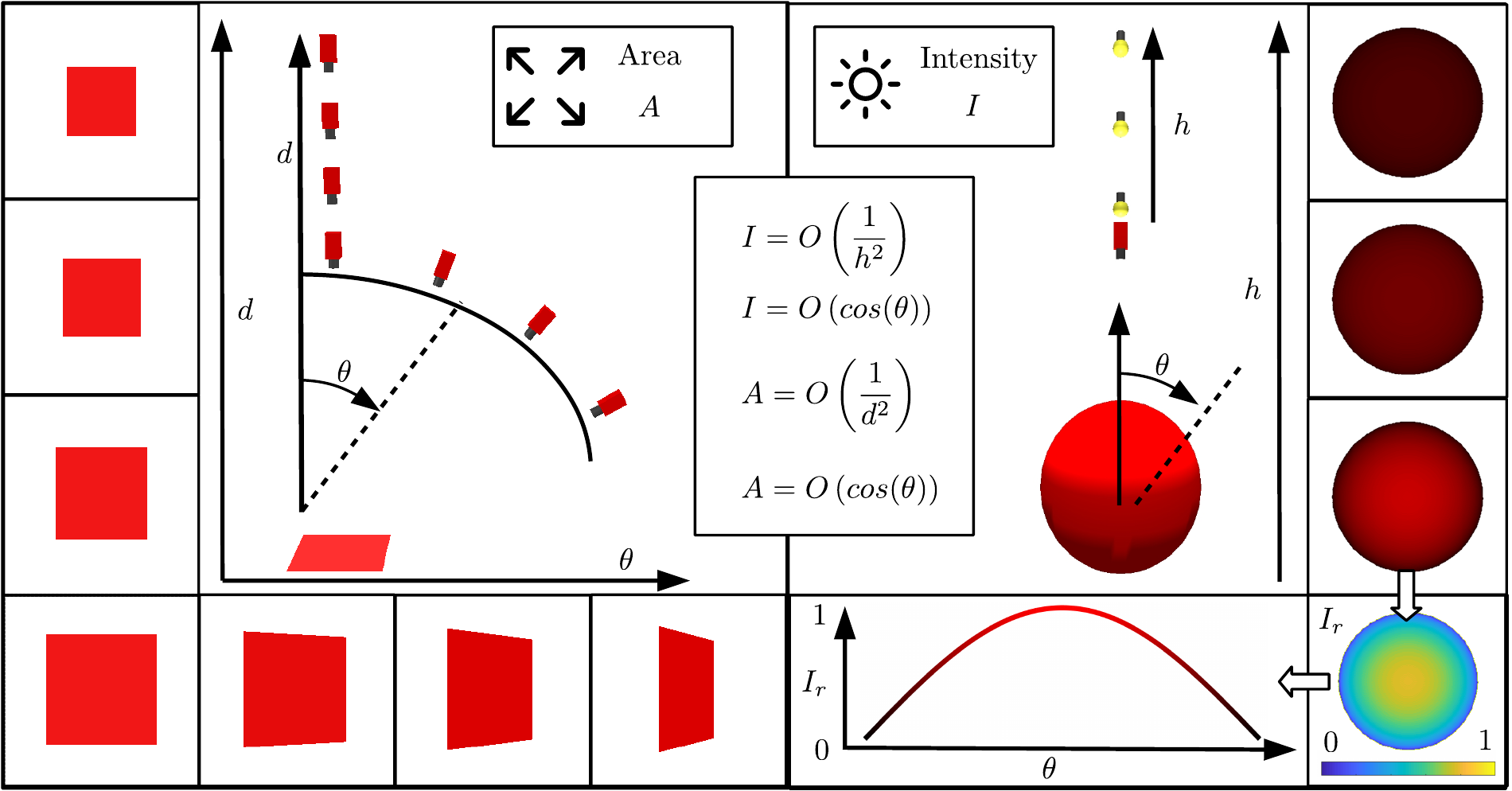}
    \caption{This image illustrates the process of constructing quality functions $Q$ from the physics of measurement processes. All the small images represent virtual images taken under respective conditions. Both the perceived area (A) and intensity (I) of features are used as examples quantities. For the intensity we assume a perfectly diffuse material. The perceived area must be large enough to detect defects during inspections. The intensity of a measurement bust be large enough to obtain a better signal to noise ratio of the measurement.}
    \label{fig:quality}
\end{figure*}

An inspection quality function can be defined as an interaction of two other functions. The first positive function $Q$ calculates for each surface point $m_i$ and each view pose $v_j$ the measurement quality (i.e. $q_{ij} = Q(m_i,v_j$). We fuse visibility information by assuming that when a surface point is invisible from a viewpoint, the measurement quality is zero (i.e. $q_{ij} = 0$). The second function fuses the quality from different measurements into one global quality value. Without loss of generality we assume that this is the $max$ function in this work. This function corresponds with taking the best measurement for each surface point during the fusion of measurement data. This makes the complete quality function:
\begin{equation}
    f(X) = \sum_{m\in M}\max_{x\in X} Q(m,x)
\end{equation}
Here $X$ can be any subset of view poses. Note that when $Q$ is binary, $f$ becomes the well-known coverage function. In \autoref{fig:quality} we illustrate how $Q$ can be constructed for physics based quality functions. In this figure we show that the perceived area of surface elements, and the image intensity, can be modelled mathematically based on measurement conditions. The size of the perceived area of a surface elements is important since smaller defects can be detected if this area is large. The image intensity is important since a brighter image results in a higher signal to noise ratio, which makes measurements more reliable. Quality functions can than be constructed by combining multiple effects that are relevant for a particular measurement technique. A more formal and complete description of inspection quality function can be found in \cite{bogaerts2019near}.

\subsection{Submodular orienteering problem}
Maximizing the inspection quality of measurements is not the only objective in the planning of inspection paths. A good inspection path must also be efficient. The problem of maximizing a submodular function with a limit on the robot's travelling budget is formally known as the submodular orienteering problem \citep{chekuri2005recursive}. In this problem the goal is to find a subset of view poses $X^*$, that maximizes the inspection quality function $f$ for a given budget $B$.
\begin{equation}
    X^*=\argmax_{X\subseteq V}\{f(X) | C(X)\leq B \}
    \label{eq:theproblem2}
\end{equation}
$C$ is a function that computes the travelling cost for any subset of view poses $X$. \citet{zhao2012automated} propose the Greedy-Cost-Benefit (GCB) algorithm to solve this problem which is near-optimal. \citet{bogaerts2019near} propose a data structure and post-processing step that ensures that near-optimal solutions can be found for large scale inspection problems using the GCB algorithm. An important advantage of this data structure is that it can also solve the inspection planning problem for robot arms. We will use this state-of-the-art algorithm (GCB) in this work as automated algorithm. This algorithm requires however that the cost function $C$ is of the following form.
\begin{equation}
    C(X) = c(X) + \alpha |X|
    \label{eq:alpha2}
\end{equation}
Here, $\alpha$ models the cost associated with performing a measurement in each view pose. $c$ is a metric distance function. This type of cost functions are however capable of representing inspection time. This is important since inspection time is the main driver of the cost of inspections, as it directly affects the object throughput. However, to compute $C$, we do not only need a discrete set of view poses $V$ but also a set of edges $E=\{e_1,...,e_l\}$ between them. The construction of this graph is quite delicate, so for its construction we refer to the original paper \cite{bogaerts2019interactive}. 

The near-optimal algorithm for the submodular orienteering problem is the greedy cost-benefit (GCB) algorithm \cite{zhang2016submodular}. This algorithms constructs a set $X$ by increasingly adding \textit{next-best elements} to the set at time $t$ (i.e. $X_t$). Note that cost $C$ increases monotonously as $X_t$ increases in size. 

\subsection{Quality comparison}
Now that the inspection planning problem is completely defined, we can derive a quality comparison that is fair. A user uses heuristics to define an inspection path, without considering the formal inspection planning problem. However, we still want to compare user generated paths with automatically generated paths. If a user generates a set of viewpoints (i.e. $X_u$) then its inspection quality can be evaluated ($f(X_u)$). However, the budget $C(X_u)$ of different paths will be different. A quality comparison that does not account for this difference is inherently unfair. Even if two paths generated by two different users are compared, it would be unfair to just compare the inspection quality without considering the travelling budget. 

For this reason we will use the idea behind $OPT$ metric introduced by \citet{bogaerts2019near}. This metric uses the fact that the solution of the greedy cost benefit (GCB) algorithm is near optimal. Thus from this solution, a relatively tight upper bound for the optimum solution can be derived. The $OPT$ metric is than the ratio of a solution and the upper bound for the optimal solution. The quality comparison we propose, calculates the budget of a solution (i.e. $C(X_u)$). After which the GCB algorithm is performed with budget $C(B)$. From this solution, the $OPT$ metric can be calculated. This normalization towards an upper bound of the optimum accounts for differences is travelling budget. But, since we also want to compare with the performance of automated algorithms, we can take the ratio of the $OPT$ metric of a solution of the automated algorithm, and the $OPT$ metric of the user generated algorithm. Since both these paths are solutions to the same abstract problem, their $OPT$ ratio is the ratio of their inspection qualities (i.e. $qr = f(X_{user})/f(X_{GCB}$). To fairly represent the power of automated algorithms, we use the $GCB+$ algorithm proposed by \citet{bogaerts2019near}, which applies a post-processing step to  improve the solution even further. 

However while we can compare the quality of inspection paths theoretically, one practical issue remains. The user is free to provide any path in the robot's workspace, while the GCB algorithm is bound to a discretization of the problem. As a result, the user generated solution $X_u$ is not a part of the workspace graph used by the automated algorithm. To account for this we include all elements of the the user's solution to the workspace graph considered by the automated algorithm (i.e. $V_+ = V\cup X_u$). We also augment the edges $E$ with all edges between elements from $V$ and $X_u$ if the distance between view poses is smaller than a predefined threshold.

\section{Virtual reality interface for robotic inspection planning}
\label{sec:interface}
The central idea behind this paper is to use virtual reality to visualize information that experts have learned to use to program inspection paths. This visualization can than help users with less experience in optical measurement techniques in programming inspection paths. The explicit visualization of this information might even help experts in the task. On the other hand, convenient user interactions should make programming robots easier for non experts. We will start by discussing the system architecture of our interface. Next, we will provide more details in two key aspects, namely the interactive quality visualization and the robot programming interaction. After this we will explicitly discuss the usability of our approach. Our implementation is publicly available in the V-REP VR Toolbox\footnote{\url{https://github.com/BorisBogaerts/V-REP-VR-Toolbox}}.

\subsection{System architecture}
\begin{figure}
    \centering
    \includegraphics[width=\columnwidth]{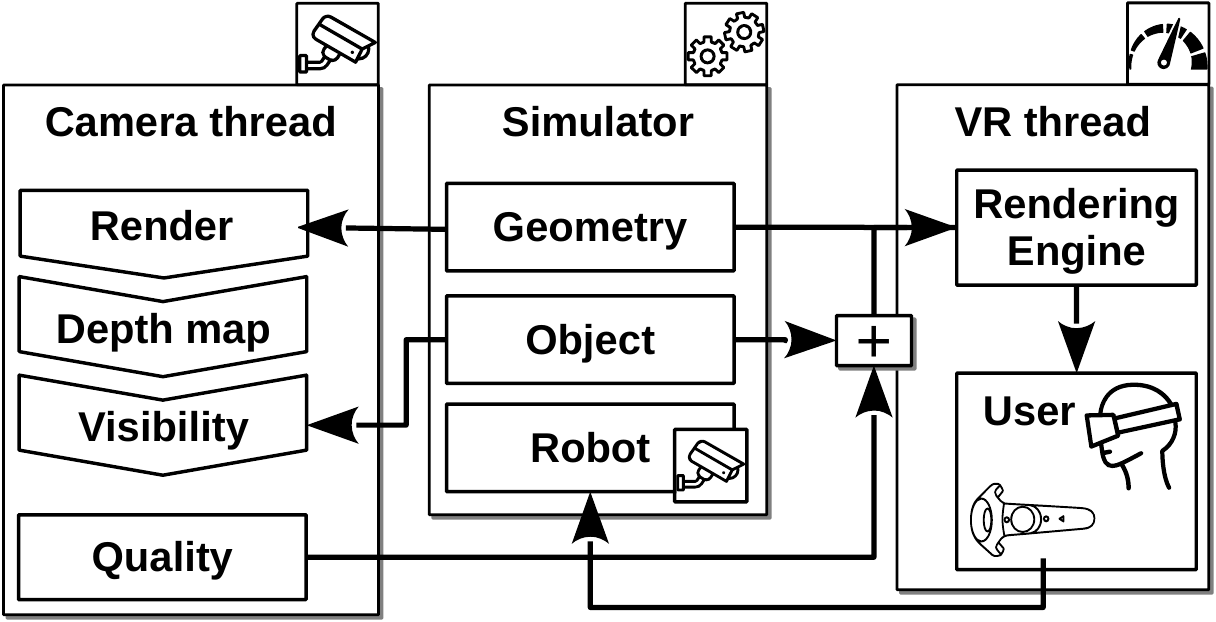}
    \caption{The main architecture of our virtual reality interface consists of three main parts. The information contained in these parts, and the interaction between them are shown in the diagram.}
    \label{fig:architecture}
\end{figure}

A schematic representation of the architecture of the interface is shown in \autoref{fig:architecture}. Our implementation consists of three main components. The central component is a robotics system simulation environment. The main virtual reality component is a virtual reality thread that renders images for the virtual reality device and returns user interactions. A third component, namely the camera thread, is necessary to create an interactive visualization of the measurement quality.

The robotic simulator is used as a modelling and physics simulation environment. In our solution we make use of a commercially available robotics simulator \cite{vrep}. The specialized nature of this simulator makes the modelling of robotic systems relatively straightforward. Objects can be imported as meshes and moved to the correct positions using a graphical user interface. This way, the user can model the robot, the measurement device and how both are connected. It is furthermore possible to model the environment in which the measurement take place, in which collisions can be predicted. The goal is to create a digital twin of the real measurement system at a usefull fidelity level. 

The VR thread renders all the content of the simulator to the virtual reality device. To create pleasant virtual reality experiences, it is necessary that fairly high framerates are achieved (60 fps minimum). Thus to ensure that we achieve such framerate the focus of this thread is rendering speed.

The architecture of our approach is such that the positions of all objects in the scene are controlled by the simulator. This ensures that all physics based calculations that are available in the simulator can be used in our approach. The VR thread simply updates the position of objects based on the position of the objects in the simulator between frames.

The camera thread is treated in \autoref{sec:visualization}. The interaction between the user and the robot system is discussed in \autoref{sec:interaction}.

\subsection{Interactive quality visualization}
\label{sec:visualization}

A previous article on camera network design, a related problem, developed a virtual reality interface that allows users to design camera networks \cite{bogaerts2019interactive}. In that work, the coverage of a camera network was visualized over the environment. Users could see how well each part of the environment was covered by the camera network, and change the position of cameras to improve this. This intuitive visualization of the abstract concept of camera coverage allowed inexperienced users to design high quality camera networks. Here, we use a similar interactive quality visualization.

\begin{figure}
    \centering
    \includegraphics[width=\columnwidth]{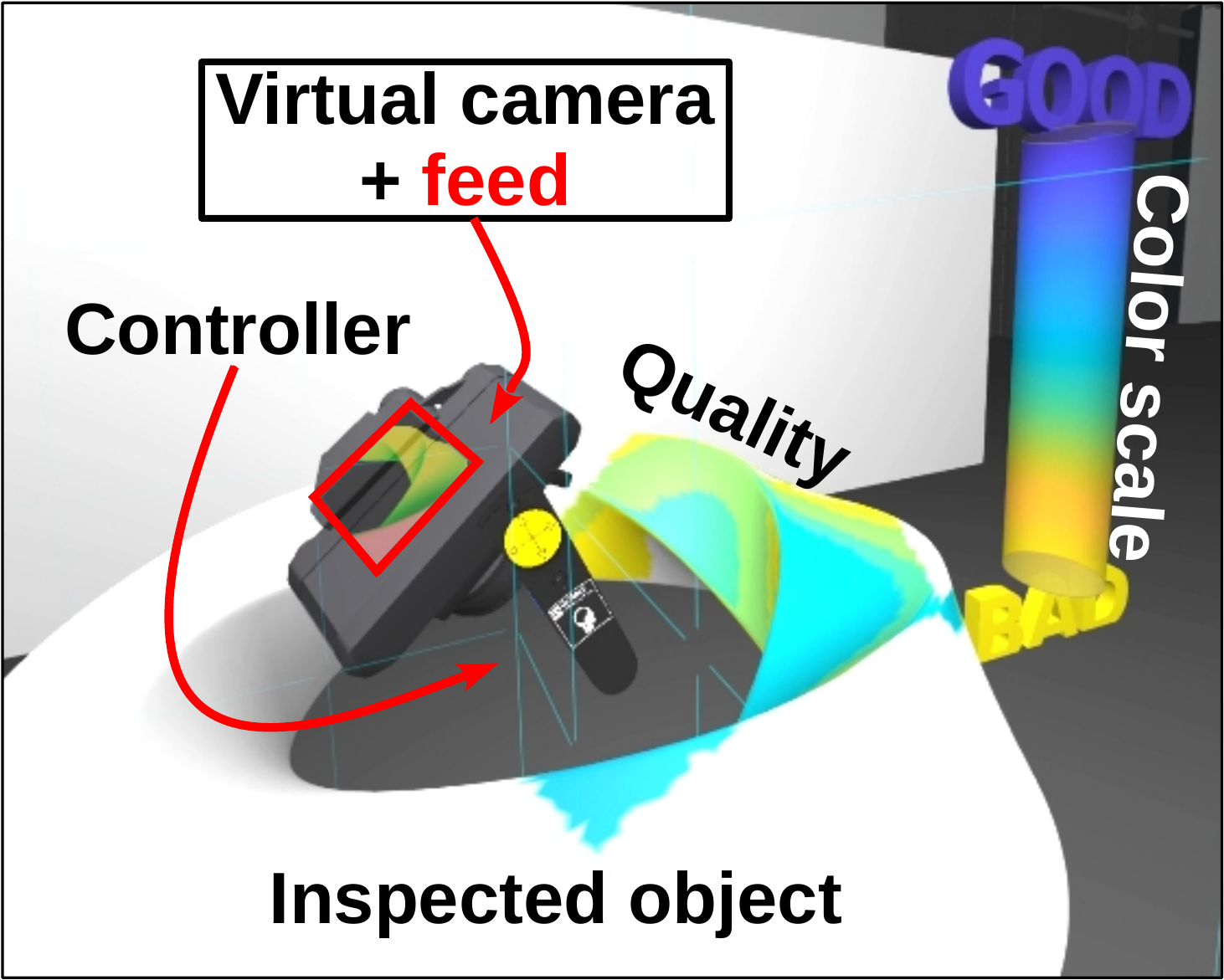}
    \caption{This image shows how users sees the interactive quality visualization. The object that is being inspected, an umbilic torus (for the enthusiasts), is dynamically colored according to the inspection quality. The scene visible in the image is also used to familiarize participants of the experiments with the concept of inspection planning. An umbilic torus was chosen as object for its interesting surface curvature which affects inspection quality. The user can move the camera which changes the colors that are being displayed.}
    \label{fig:qualVis}
\end{figure}

The basic idea of the coverage visualization is simple, the current inspection quality is visualized on the surface of the object that needs to be inspected. The user can move the virtual camera after which the displayed quality changes. This is shown in \autoref{fig:qualVis}. If a camera is located in position $x_t$, the value of $q_{m,t} = Q(m,x_t)$ is visualized at each point of the object's surface. The user can also record a trajectory after which the value is updated to $q_{m,t} = max \left(q_{m,t-1}, Q(m,x_t) \right)$. This visualizations shows the user how well different parts of the object are measured, and which parts of the objects are not measured at all. 

The interactive quality computation is performed in the camera thread in our implementation. In this thread, all environment objects that are defined in the simulator, are rendered from the current perspective of the measurement device (i.e. $x_t$). From this render, a depth map is extracted. The visibility of points $m \in M$ can be determined by comparing their depth values with the depth values stored in the depth map. All the points that have a depth value which is smaller (or equal) to the depth value of the depth map are visible. This algorithm to compute the visibility of points is known as Z-buffer visibility computations. This approach was favored over alternatives, such as ray-tracing to guarantee a real-time (i.e. interactive) computations. 

We furthermore compute quality values $q_{m,t}$ in OpenGL shaders during the rendering process. After the visibility calculation and calculation of quality values it is straightforward to determine the inspection quality. The values $q_m$ are attached to each mesh vertex $m\in M$. The VR thread can than efficiently interpolate the vertex data over each triangles using standard OpenGL shaders. The fact that most of the computations are performed in shaders ensures that the quality values are updated in real-time. 

\subsection{Robot programming interaction}
\label{sec:interaction}
Controlling the robot should be as intuitive as possible, as no user experience can be presumed. To achieve this, the user can directly control the Tool Center Point (TCP) of the robot with a virtual reality controller. Since the three dimensional position and orientation of the virtual reality controller are tracked, it is possible to completely control the TCP of the robot. This approach also eliminates the need for advanced orientation control methods \cite{orientation}.

\begin{figure}
    \centering
    \includegraphics[width=\columnwidth]{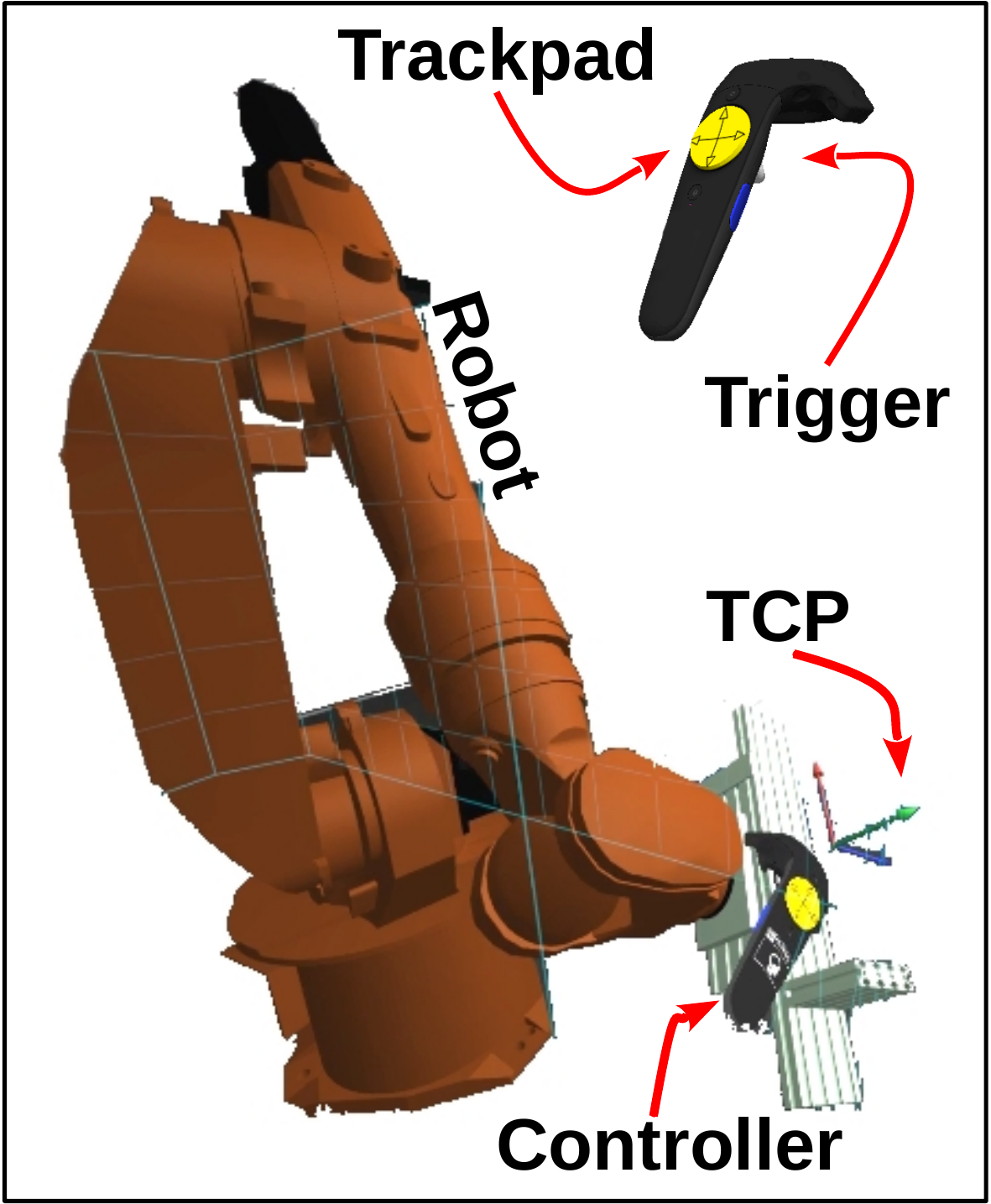}
    \caption{The user can control the TCP of the robot by pressing a trigger button on the controller, after which the robot follows the controller. The position of the trackpad and trigger buttons are also shown. These buttons are color coded during the experiments to effectively communicate with the user.}
    \label{fig:robotInter}
\end{figure}

We avoid however that the robot follows the controller the entire time. The controller can move freely until the user presses the trigger button, after which the transformation between the TCP and the controller remains fixed, even if the user moves the controller. Releasing the trigger button releases the TCP. As a result the robot follows the user's controller only if the trigger button is pressed. The position of the buttons on the controller are shown in \autoref{fig:robotInter}. This feature is important since the user can choose the optimal position relative to the robot before controlling it, especially since occlusion of the inspected object by the robot from the perspective of the user must be avoided.

The robot follows the TCP by continuously performing inverse kinematics calculations. To ensure that the motion of the robot is continuous we use a damped least squares solver. Joint angles found by the inverse kinematics solver are only applied to the robot if the inverse kinematics calculation is successfully. So if users move the TCP target out of reach of the robot, the robot simply stops moving. The user can use this visual cue to reconsider movement of the TCP to positions that are reacheable.

The user can record trajectories by pressing the trigger button of the second controller. This recorded path is also visualized to the user by a simple line in 3D. Internally, the joint angles of the path are stored. When the TCP is released by the main controller, the user can scroll through the part of the path that already had been recorded by scrolling over the trackpad. The previously recorded joint angles are then re-applied to the robot in the reverse order in which they were recorded. 

Some experiments feature a robot which contains an extra axis. This can be a rotation table (see \autoref{fig:objectsAndRobots} (middle) for a robot with rotation table), or a linear translation stage (see \autoref{fig:expPrep} under `the Challenge' for a robot with translation stage). These robot systems cannot be controlled by simply moving the TCP. In these cases, the user can control this extra axis by scrolling over the trackpad of the second controller.

Note that in any sensible robotic inspection system, the measurement device is attached to the robot end-effector. Thus any movement of the robot's TCP results in a motion of the measurement device. Thus after moving the robot's TCP, the interactive quality visualization re-calculates the inspection quality.

\subsection{Usability}
\label{sec:usability}
In this section we will reflect on the some fundamental issues in the usability of traditional workflows with automated algorithms, that can be solved by replacing automated algorithms with humans.

A fundamental issue in automated inspection planning is related to the dimensionality of the state space of the robot system. In reality, high dimensional state space result in robots that are more flexible, and thus have more path options, of which some can be better that what is achievable with a less flexible robot. Automated algorithms do not deal well with high dimensional state spaces. This is because the inspection planning problem is fundamentally characterized by a long planning horizon \cite{binney2012branch}. Searching over longer time horizons in larger state spaces drastically reduces the search area that can be covered by automated algorithms (due to an exponentially increased computational complexity \cite{zhang2016submodular}). Thus with automated algorithms, the performance on robots with larger state spaces typically deteriorate. So, at least partially, circumvent these issues complicated data-structures are required \cite{bogaerts2019near}. These data structures are delicate to set up, and typically require user expertise. On the other hand, users with our interface will run into less issues when searching for inspection paths with highly flexible robots. The effect will typically be that robots are easier to control by users.

While digital twins aim to be perfect representations of real world systems, they can still fail in certain respects. For inspection planning this can be especially the case since robotic inspection systems are complex mechatronic systems. On the other hand, since the inspection planning problem is challenging digital twin fidelity is often sacrificed for an increase in simulation speed \cite{peeters2019optimized}. An example that can cause such problems are cables from the measurement setup that get tangled. Cable movement and tangling can be simulated, however their simulation will in many cases be too expensive, and get therefore omitted. If such a thing happens, it requires an expert to fix the digital twin before automated algorithms can re-solve the problem. In our approach, the user, who does not need to be an expert, can simply take this aspect into account when programming a trajectory. 

Another fundamental issue with automated algorithms is the fact that user preferences are often challenging to encode in inspection quality functions. In previous work, it has been shown that in some cases these functions only partially reflect user preferences \cite{bogaerts2019interactive}. While the visualization, that the user gets to see, does not visualize some unmodelled aspects, users who are familiar with these preferences still take them into account \cite{bogaerts2019interactive}. Automated algorithms, on the other hand, rely on the success of experts in encoding these challenging aspects.

\section{Experiments design}
\label{sec:experimentDesign}
The experiments consist of a user study, in which participants of different experience levels in robotics and optical inspection techniques generate inspection paths for different inspection cases. Our user study consists of three main stages\footnote{A 360VR video that provides an overview of our experimental procedure is available online \url{https://youtu.be/pFAptrCYhaQ}. The VR format helps to reproduce the experience of the experiment from the perspective of the participant.}. In the first stage, participants are introduced to all visualizations and interactions that can be performed with the virtual reality interface. In the second stage, the participants generate inspection paths for six different inspection planning scenarios. And in the last optional stage, the users are asked to generate an inspection path for a more complicated and larger scale inspection task. This task is significantly more challenging and requires more effort of the user, and is therefore optional. The user study is designed such that the first two stages can be completed in approximately one hour, and the last optional case adds up to an half hour to this. However, users are never pressured with time limits, and inspection paths are only stored if the user is pleased with them. Furthermore, users are reminded that they can stop the study at any point if they do not wish to continue.

\subsection{User selection}
An important contribution of this work is to investigate the importance of the experience level of users in either robotics, inspection techniques or both. The virtual reality interface aims to compensate a lack of experience by providing an intuitive understanding that is sufficient to successfully complete the inspection planning task. To investigate this, a population of participants with a range of experience levels in both robotics, inspection techniques or both are required. In order to find such specific profiles we targeted specific people. This specific targeting of users was deemed necessary to obtain a decent variation of participants with varying experience levels. To measure the experience level of the participants, we asked each participant to rate their experience in both domains by a number from one to five according to following descriptions:
\begin{itemize}
\item 1 - I do not have any experience in the field
\item 2 - I have only theoretical knowledge about the field, but no practical experience
\item 3 - I have programmed a robot once or twice/ I have performed an optical measurement once or twice.
\item 4 - I program a robot arm regularly/ I perform optical measurements regularly
\item 5 - I have experience in designing robot systems/ I have experience in planning optical inspections 
\end{itemize}
we performed the experiments with 16 participants. The participants included both males and females in an age range of 22-58. The distribution over all the combinations of experience levels is shown in \autoref{fig:result1}.

\subsection{User preparation}
\begin{figure*}
    \centering
    \includegraphics[width=\textwidth]{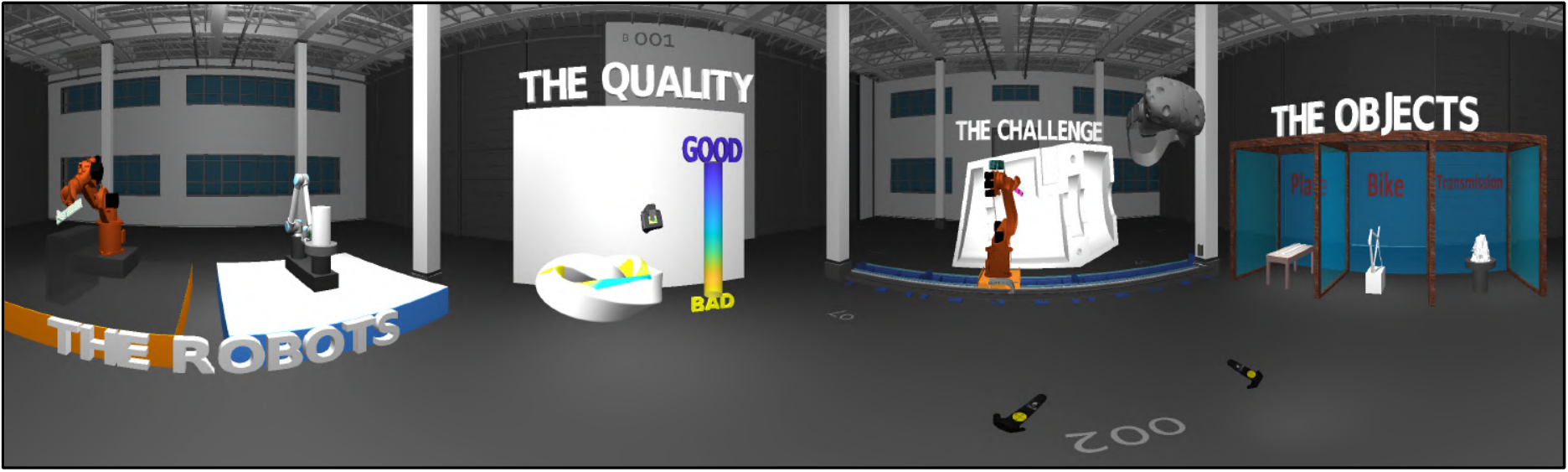}
    \caption{360 degree image showing the virtual reality scene that is used to prepare participants for the experiments. This scene consists of three main parts. 'The quality' familiarizes users with the concept of inspection quality and how it is visualized. In 'The objects' users get to see the objects that need to be inspected in the experiments. In 'The robots' users learn how robots can be controlled.
    }
    \label{fig:expPrep}
\end{figure*}
In this first step of the user study, users are familiarized with the controls, the inspection planning problem, inspection quality visualization and robotics controls. To prepare the participant, a special scene was created which consists of three main elements. In the first element named `the quality' (see \autoref{fig:expPrep}) users are familiarized with the concept of inspection quality. The users learn how they can move a camera, and see the inspection quality interactively projected onto a test object. In this step users can freely move the camera and see the effect on the inspection quality. Participants also learn how to record measurement trajectories. 

Participants are transferred to the second step, which is called `the objects' (see \autoref{fig:expPrep}), if they feel familiar with the concept of inspection quality. In this step, users can study the objects for which they need to generate inspection paths in the following stage of the experiments. The user can already try to find inspection paths for these objects, without the constraints imposed by a robot system. To do this they use the same controls from the first step, and by doing so get more intuition about the inspection quality. 

Finally, the users are transferred to the last step, named `the robots', to learn how robot paths can be programmed with the VR interface. The participants learn this in two different scenarios. In the first scenario users learn how to move the end effector of the robot, a Kuka kr16 which is also used in further experiments (see \autoref{fig:objectsAndRobots}), to a predefined set of configurations. These predefined set of poses are chosen such that the users experiences collisions and the kinematic limitations of the robot. Each time such a collision or kinematic limitation happens, users are informed about this, and it is explained that this should be avoided during the experiments. In the second scenario, users learn how robot paths can be programmed. This time, the robot is a UR10 with rotation table (see \autoref{fig:objectsAndRobots}), as is used in further experiments. Two predefined trajectories are one by one shown, which need to be programmed by the user. The user thus needs to record a trajectory while following the shown trajectory with the robot end-effector. One trajectory is a helix which can be programmed by rotating the rotation table, and moving the Tool Center Point (TCP) of the robot linearly. This path was chosen to show users how a rotation table can be used effectively in combination with a robot arm.

This user preparation step also aims to guard the experiment against users that have difficulties with navigating in, and interacting with virtual environments in Virtual Reality (VR). As experience with VR is not a factor of interest in this work, it could taint the results. To avoid this, users that cannot complete the user preparation cannot proceed to the following stages. From the 16 participants that entered the study, only one participant was rejected as a result from this preparation step. Finally, the user is presented with 'the challenge', which is the third optional stage of the experiment.    

\subsection{Inspection planning scenarios}
\begin{figure*}
    \centering
    \includegraphics[width=\textwidth]{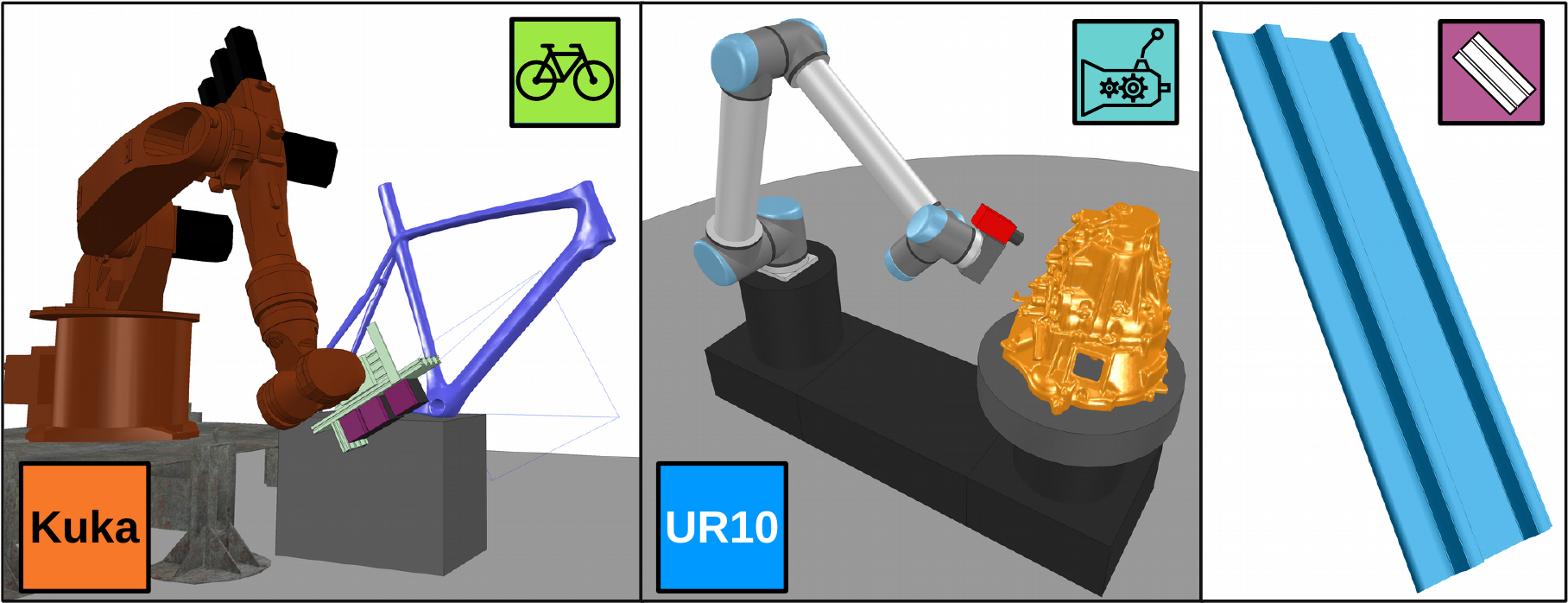}
    \caption{The tree different objects, and two different robots systems that are featured in the experiments. All possible combinations between robots an measurement objects make up the six cases that the users must solve.}
    \label{fig:objectsAndRobots}
\end{figure*}
The next stage in the experiments is the most important part. Participants are asked to program inspection paths that are as efficient as possible in six different cases. These cases are constructed by combining two different robot systems with three different objects. These are shown in \autoref{fig:objectsAndRobots}. During these experiments each user is free to try to program as many trajectories as desired. The user can stop each case if he is pleased with the final measurement trajectory. The user can also ask to store intermediate trajectories, which can be considered as final trajectory.

The three different objects that are featured in the experimental evaluation are a bicycle frame, a transmission housing and a panel. The panel is the simplest object, for which it is relatively straightforward to find a good inspection path. However, the slopes in this object make it slightly more interesting. The transmission housing is a nearly convex object, with a lot of small scale occlusions. The third object is a bicycle frame, for which the most complicated inspection path is required. Furthermore, two robots are considered in the experiments. The first is a Kuka kr16 robot with a relatively large measurement device, which is the simplest robot system to control. The second robot is a more flexible UR10 together with a rotation stage. The latter is a more complicated robot system because both the robot's TCP, and the rotation stage must be controlled separately by the user. 

\subsection{Optional challenging inspection scenario}
In the final stage of the user study, users need to program a more challenging inspection path. This scenario is displayed in \autoref{fig:problems2}. The robot during this scenario is a Kuka kr16 that is placed on rails. The user can control this extra degree of freedom with the left controller trackpad. This is the same control that is used to control the rotation stage in previous experiment. The object that is being inspected is the housing of a wind turbine. This is a large and challenging object. Another challenging aspect of this object are deep notches, in which the robot must navigate to obtain good measurements. This experiment is optional because it requires much concentration of the participant, which could be lacking after previous stages of the study. Surprisingly, all participants agreed to perform this extra step. 

\subsection{Inspection problem definition}
\begin{figure}
    \centering
    \includegraphics[width=0.4\textwidth]{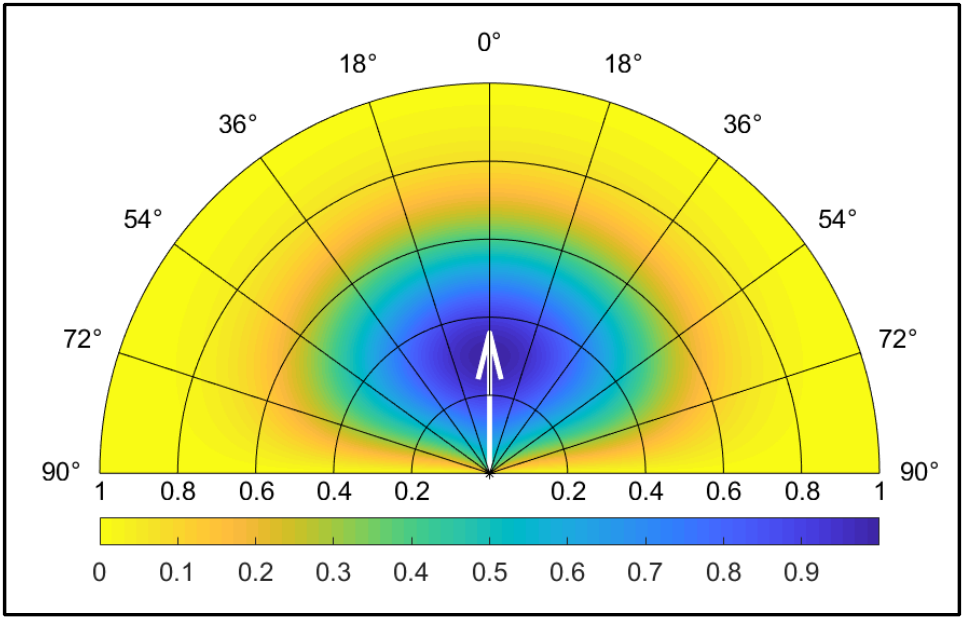}
    \caption{The inspection quality function used throughout the experiments. The white arrow represents a surface normal. The color represent the inspection quality if a camera is located in a position, and points towards the base of the white arrow. A higher number of the inspection quality indicates a better measurement.}
    \label{fig:qualityFunction}
\end{figure}
In this section, we will give some details on how all the functions that define the submodular orienteering problem are chosen in the experiments. We assume throughout the experiments that the inspection quality function $Q$ is of the following form:
\begin{equation}
    Q(d,\theta) = \cos(\theta) \cdot \exp{\left ( \frac{(d-d_{opt})^2}{\sigma^2} \right )}
\end{equation}
Here, $\theta$ is the angle between the surface normal and the view direction of the camera, and $d$ is the distance between a point and the viewpoint of the camera. The cosine dependence on the scanning angle is a common aspect of quality function, as is clear from \autoref{fig:quality}. $d_{opt}$ and is the ideal measurement distance, and $\sigma$ is used to determine how much the measurements can deviate from this ideal distance. 

To define the distance function $C$ (\ref{eq:alpha2}) we assume that $\alpha$ is zero. This is because we focus on continuous measurements. We further assume that function $c$, which considers edge costs of the graph is of the following form:
\begin{equation}
c(e_{i,j})=(1-\beta)c_t(p_i,p_j)+\beta c_o(o_i,o_j)
\label{eq:dist2}
\end{equation}
Here $c_t$ is the euclidean distance between positions connected by an edge, and $c_o$ is the angle in radians of the axis angle rotation between orientations. To complete the definition of this function we assume that $\beta$ is 0.01. 

\section{Experimental results}
\label{sec:experimentResults}

\subsection{Small scale inspection planning problems}
In all the inspection cases we define $d_{opt}$ as 200mm, and $\sigma$ as 100mm. In each case, the automated algorithm has a time limit of 30 minutes to compute an inspection path. 
\begin{figure*}
    \centering
    \includegraphics[width=\textwidth]{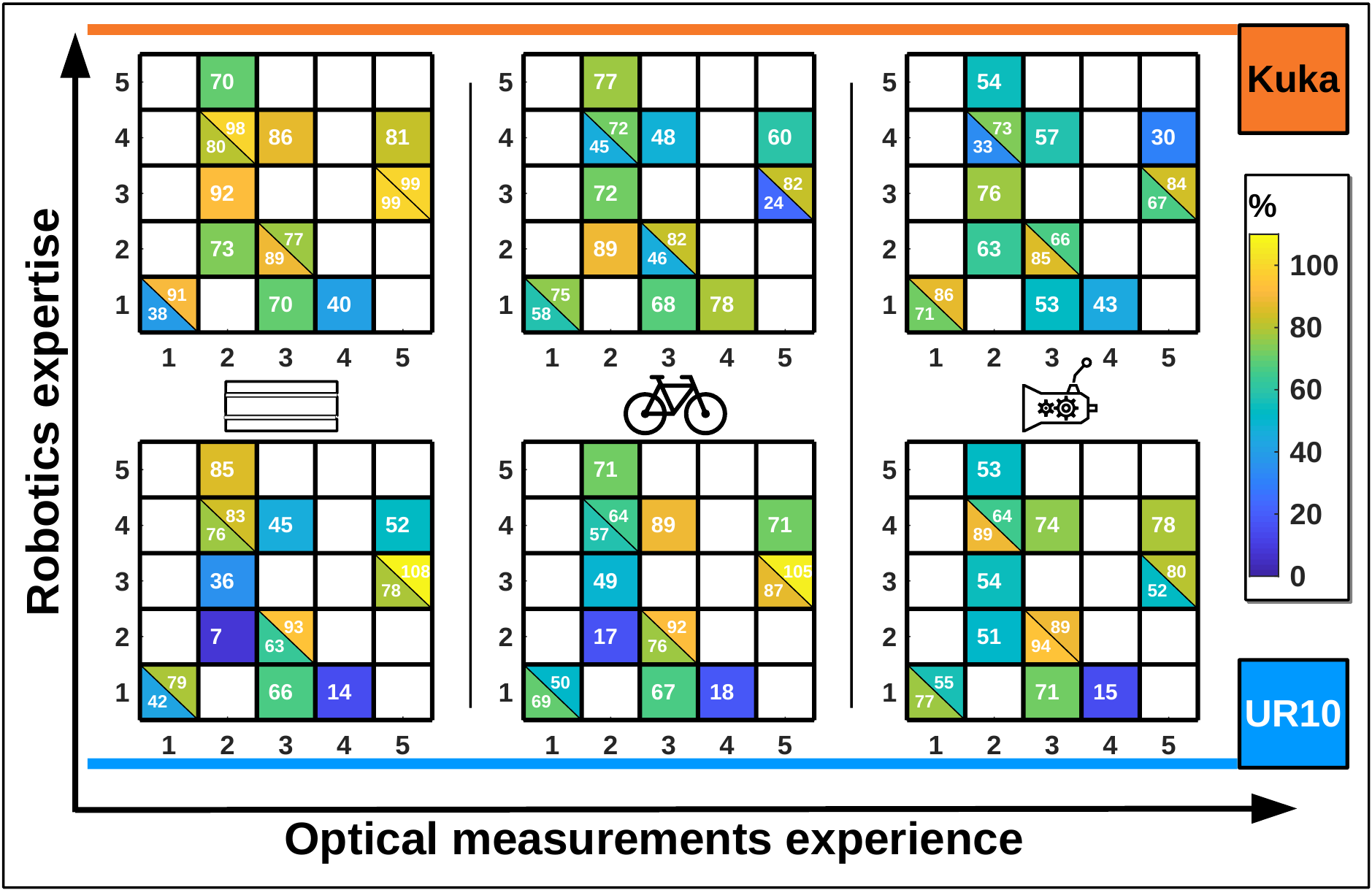}
    \caption{This figure shows the ratio of the inspection quality of user generated paths, and automatically generated paths with the same budget. This ratio is given for each user, and inspection scenario in the second phase of the experiment. Sometimes, a square in the experience level matrix is subdivided in two. This occurs when two users have the same experience level.}
    \label{fig:result1}
\end{figure*}
The results of the six small scale experiments are provided in \autoref{fig:result1}. These results show that there is a large range in user performance. But on the other hand it is also notable that users with very little experience can provide inspection paths that are good compared to specialized automated algorithms. The lowest median user performance (lowest of different inspection scenarios) is 66\%. There were only three cases in which six participants scored under 60\%. These are: kuka-transmission, UR10-panel and UR10-transmission. It is notable that users scored badly on the UR10-panel case and better with the UR10-bike case, while the opposite happens in the Kuka cases. This indicates that this dip in user performance may be due to the switch of robots, which happened just before the UR10-panel case. It also appears that users have trouble with the transmission object. The group of users that scored over 60\% always included users with little to no experience.

The time that users needed to find an inspection path ranged from 21 seconds to 8 minutes and 20 seconds. In this maximum time, the user generated a few inspection paths. Thus each individual inspection path was generated in a shorter time.
\begin{figure}
    \centering
    \includegraphics[width=0.5\textwidth]{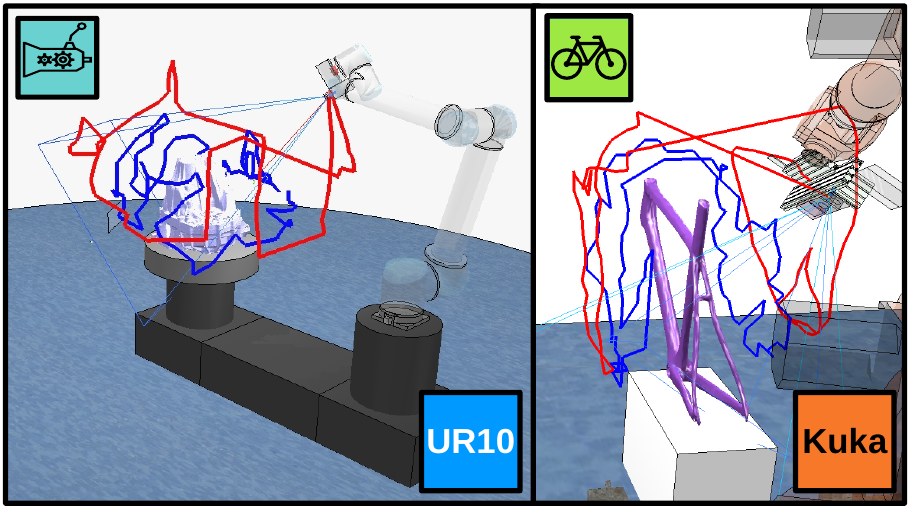}
    \caption{These images show two cases where users obtained a very low quality ratio. The path generated by the user is shown in red, and the path generated by the automated algorithm is shown in blue. On the left a user (1-robotics, 4-inspection) achieved a quality ratio of 15\% on the turbine-UR10 inspection scenario. On the right a user (3-robotics, 5-inspection) achieved a quality ratio of 24\% on the bike-Kuka kr16 inspection scenario.}
    \label{fig:problems1}
\end{figure}
A visual analysis of the lower quality inspection paths generated shows that these paths typically contain some strategical mistakes. Two examples are shown in \autoref{fig:problems1}. In \autoref{fig:problems1} (left) a user generated an inspection path with a large radius. As a result, the inspection budget increases rapidly. On the other hand, the automated algorithm generated a path with a smaller radius. This path is at a distance that is optimal according to the inspection quality function. Thus the choice to generate a larger radius inspection path is punished twice. On one hand, the inspection quality is lower, and on the other hand the budget is larger. This effect may explain the lower performance of users with the transmission object. In \autoref{fig:problems1} (right) the user generated an inspection path with a radius that was too large. But the inspection path also contains a few detours. In these extra detours, budget is spent without a notable gain in inspection quality.  ddd

\subsection{Large scale inspection planning problem}
In this inspection case we define $d_{opt}$ as 300mm, and $\sigma$ as 300mm. The automated algorithm has a time limit of 5 hours. 
\begin{figure}
    \centering
    \includegraphics[width=0.5\textwidth]{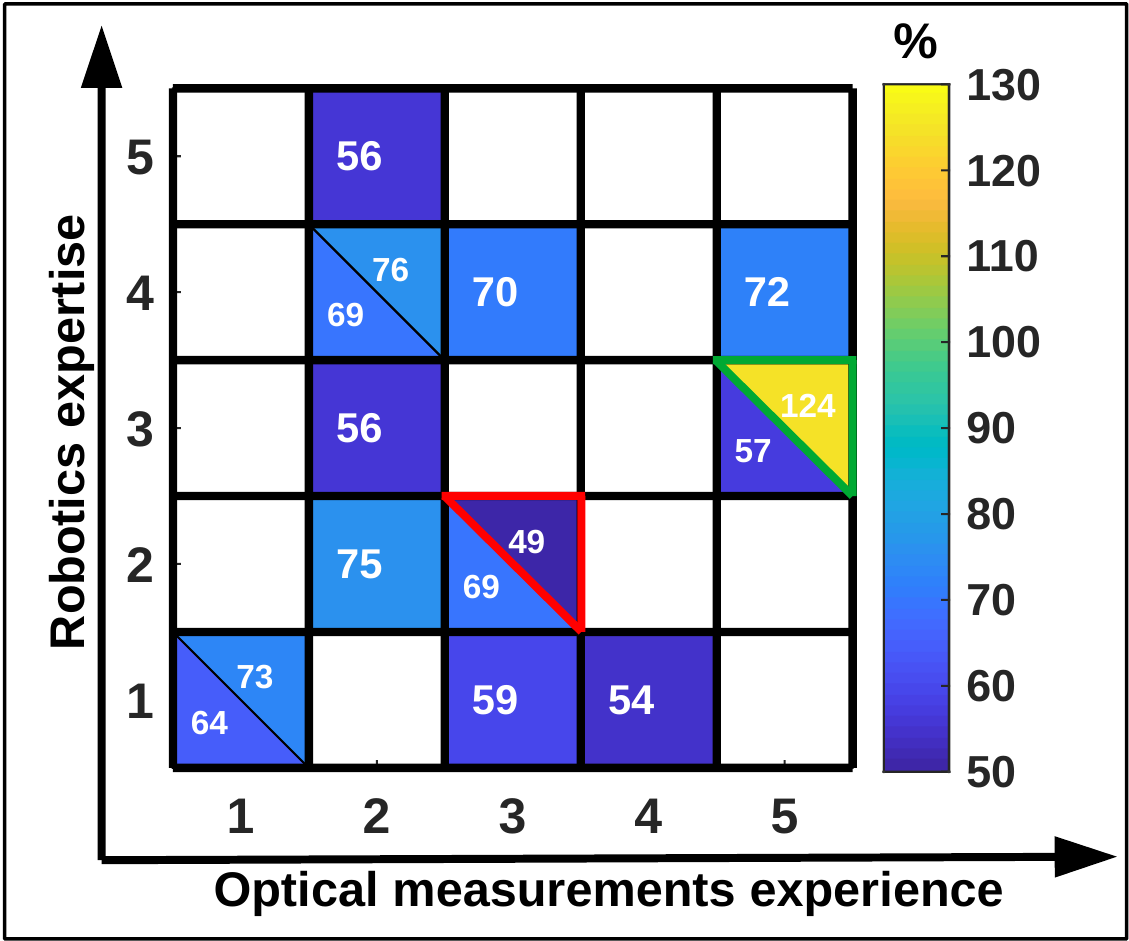}
    \caption{This figure shows the ratio of the inspection quality of user generated paths, and automatically generated paths with the same budget. This ratio is given for each user. Each user that entered the experiments performed this optional step voluntarily. Notice that the low end of the colorscale is 50\% compared to the low end of the colorscale of 0\% in \autoref{fig:result1}. The inspection paths of the triangles with red an green edges are shown in \autoref{fig:problems2}.
    }
    \label{fig:result2}
\end{figure}
The result of the large scale, optional inspection planning scenario are provided in \autoref{fig:result2}. It is notable that the lowest inspection quality is 49\%, which is significantly better than in previous experiments. And users with little experience in either robotics or inspections, are again able to generate decent inspection paths. The median performance of all users is 69\%. Notable is that a user managed to achieve to generate a path with a 24\% higher inspection quality than the automated algorithm with the same budget. It is furthermore notable that there are less outliers with a bad quality than with the smaller scale inspection problems. This is mainly because the longer length of the total inspection path. The effect of an inefficient detour is less for longer paths. The higher inspection quality of user generated paths indicate that the negative effect of detours decreases with longer inspection paths. The time that users needed to generate an inspection path ranged from three minutes and 29 seconds to 9 minutes and 37 seconds. 
\begin{figure*}
    \centering
    \includegraphics[width=\textwidth]{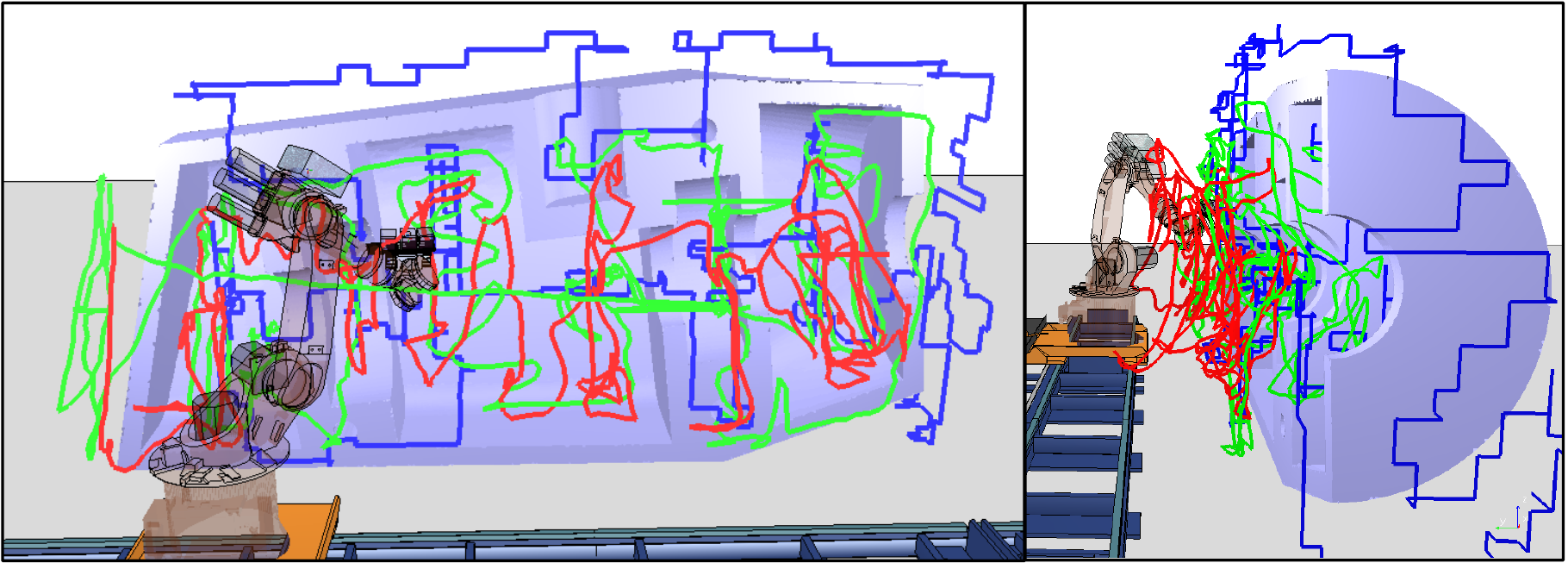}
    \caption{This figure shows an inspection path generated by the automated algorithm (blue), the best user generated inspection path (green) and the worst user generated inspection path (red). 
    }
    \label{fig:problems2}
\end{figure*}
In \autoref{fig:problems2} three inspection paths are shown. The best (green) and worst (red) user generated paths are shown, as well as the computer generated path (blue). The red inspection path contains the strategical disadvantage that it is further from the object, which leads to a worse inspection path. It is furthermore more compact, which leads to a lower total coverage. Compared to the green and blue paths it also contains significantly more detours. 

\subsection{Discussion}
\label{sec:Discurssion}

A first important side note is that during the user study, users were barely educated in using our interface to define inspection paths. In fact, all users finished the study within an hour. Considering that users learned how to use the interface, generated six smaller scale inspection paths and one complex inspection path, this is a short time frame. Users even reported that their performance would likely improve with more experience. Users also reported that if they made more attempts, their inspection paths would also likely improve. Thus the results that were achieved in this user study provide lower bounds on the quality of inspection paths that well trained users can provide. This limited education of users is also partly responsible for the large variation in user performance.

The experiments also showed that when quality and efficiency are key, user generated paths are not yet an option. In most cases, user generated paths have a significantly lower quality than computer generated ones. We identified some strategical areas in which user generated algorithms can improve that can be targeted by future approaches: 
\begin{enumerate}
    \item Fix positioning inaccuracy
    \item Managing robot complexity
\end{enumerate}
During the experiments it was clear that the positioning inaccuracy of users has a big impact on inspection paths. One result of this is that users tended to position the measurement device at a sub-optimal measurement distance. As a result, measurement paths become longer, and inspection quality decreased. Another side effect of this inaccuracy is that it resulted in detours. When a user missed some spot of an object (which could have been covered given a more delicate positioning) users tended to fix this by generating a detour. This detour resulted in two parts of the inspection path that were close to eachother. This again, increases the length of measurement paths without adding much to the inspection quality. Another complexity for users was related to the complexity of controlling robot systems. Sometimes users got distracted by managing the robot, which generates detours. 
 
An interesting direction for future research is to investigate whether these problems can be solved in either post-processing, or with on-line aiding systems. In path processing either local \cite{bogaerts2018gradient} or global \cite{yi2014informative,reardon2018shaping} solutions are possible. Human aiding systems can focus on decreasing the degrees of freedom that must be controlled by the user. An example of a degree-of-freedom that is better not controlled by a user is the distance of the measurement device to the object.

\section{Conclusion}

In this work, we investigated if inexperienced users can be used to generate high quality inspection paths. The main idea of our solution is to replace the need for specialized experience, by intuitive visualizations and interactions in virtual reality. To quantify the performance of user generated inspection paths, we proposed an approach based on the abstract structure behind the inspection planning problem. In this approach user generated inspection paths are compared with inspection paths resulting from a near-optimal inspection planning algorithm.

To investigate if specialized experience can be replaced by an intuitive interface we performed a user study with 15 valid users with different experience levels in robotics and optical inspections. In this user study users solved six smaller scale, and one large scale inspection problem. From this user study it was clear that, while user performance was variable, users without experience could generate high quality inspection paths. The median user performance was in the range of 66-81\% of the quality of a state-of-the-art automated algorithm. We also discussed the source of this variability in user performance and how this could be solved in future.

From our experiments it is clear that users were able to generate inspection paths much faster than automated algorithms. Especially in the complex inspection planning scenario, users needed maximally 9 minutes and 37 seconds to generate an inspection path, while the automated algorithm needed 5 hours. A human-centric approach to inspection planning is also easier to set up, since digital twin fidelity is less important. Furthermore, the notion of inspection quality does not need to be defined as precisely. It only serves to give users an indication of inspection quality.

\IEEEpeerreviewmaketitle

\section*{Acknowledgment}
B.B was funded by Fonds Wetenschappelijk Onderzoek (FWO, Research-Foundation Flanders) under Doctoral (PhD) grant strategic basic research (SB) 1S26216N.

\ifCLASSOPTIONcaptionsoff
  \newpage
\fi

\bibliographystyle{IEEEtranN}
\bibliography{references}



\end{document}